\documentclass{article}

 \usepackage[main, final]{neurips_2026}

\usepackage[utf8]{inputenc} 
\usepackage[T1]{fontenc}    
\usepackage{hyperref}       
\usepackage{url}            
\usepackage{booktabs}       
\usepackage{amsfonts}       
\usepackage{nicefrac}       
\usepackage{microtype}      
\usepackage{xcolor}         
\usepackage{amsmath}
\usepackage{multirow}
\usepackage{pifont}

\usepackage{graphicx}
\usepackage{colortbl}
\usepackage{makecell}
\usepackage{wrapfig}
\usepackage{caption}

\title{Bridging Scene Generation and Planning: \\
Driving with World Model via Unifying Vision and Motion Representation}

\author{
Xingtai~Gui\textsuperscript{\rm1},\hspace{0.75em}
Meijie~Zhang\textsuperscript{\rm2},\hspace{0.75em}
Tianyi~Yan\textsuperscript{\rm1},\hspace{0.75em}
Wencheng~Han\textsuperscript{\rm3}\\
\textbf{
Jiahao~Gong\textsuperscript{\rm2},\hspace{0.75em}
Feiyang~Tan\textsuperscript{\rm2},\hspace{0.75em}
Cheng-zhong~Xu\textsuperscript{\rm1},\hspace{0.75em}
Jianbing~Shen\textsuperscript{\rm1}\thanks{Corresponding author. This work was supported by the Science and Technology Development Fund of Macau SAR (FDCT) under grants 0134/2025/RIA2 and 0102/2023/RIA2, and the Jiangyin Hi-tech Industrial Development Zone under the Taihu Innovation Scheme (EF2025-00003-SKL-IOTSC).}}
\\[0.5em]
\textsuperscript{\rm 1} SKL-IOTSC, Department of AI, University of Macau \\
\textsuperscript{\rm 2} Afari Intelligent Drive,\hspace{0.75em}
\textsuperscript{\rm 3} School of Control Science and Engineering, Shandong University
}

\begin{document}

\maketitle

\begin{abstract}
End-to-end autonomous driving aims to generate safe and plausible planning policies from raw sensor inputs, and constructing an effective scene representation is a critical challenge. Driving world models have shown great potential in learning rich representations by predicting the future evolution of a driving scene. However, existing driving world models primarily focus on visual scene representation, and motion representation is not explicitly designed to be planner-shared and inheritable, leaving a schism between the optimization of visual scene generation and the requirements of precise motion planning. We present WorldDrive, a holistic framework that couples scene generation and real-time planning via unifying vision and motion representation. We first introduce a Trajectory-aware Driving World Model, which conditions on a trajectory vocabulary to enforce consistency between visual dynamics and motion intentions, enabling the generation of diverse and plausible future scenes conditioned on a specific trajectory. We transfer the vision and motion encoders to a downstream Multi-modal Planner, ensuring the driving policy operates on mature representations pre-optimized by scene generation. A simple interaction between motion representation, visual representation, and ego status can generate high-quality, multi-modal trajectories. Furthermore, to exploit the world model’s foresight, we propose a Future-aware Rewarder, which distills future latent representation from the frozen world model to evaluate and select optimal trajectories in real-time. Extensive experiments demonstrate that WorldDrive achieves leading planning performance among vision-only methods while maintaining high-fidelity action-controlled video generation capabilities. Code is available at \url{https://github.com/TabGuigui/WorldDrive}.

\end{abstract}

\section{Introduction}
\label{sec:intro}
End-to-end autonomous driving aspires to learn direct sensor-to-action policies~\cite{review1, review2, review3}, which hinges on effective visual representation learning~\cite{uniad, sparsedrive, zheng2024occworld, driveworld}. Propelled by recent breakthroughs in generative modeling, Driving World Models~(DWMs) are emerging as a promising paradigm for autonomous driving~\cite{review4, review5, review8}. By explicitly modeling the scene evolution, DWMs provide a powerful foundation for forecasting complex driving environments.

Leveraging this predictive capability, the representation learned by DWMs holds immense potential for downstream planning. Current integration strategies generally follow two main paradigms. One approach leverages high-fidelity models to generate future scenes for downstream planners~\cite{drivewm, zeng2025fsdrive, zhao2025pwm, li2025imagidrive}. However, this process incurs prohibitive computational costs. To mitigate this overhead, a second paradigm operates entirely within a latent world model~\cite{law, world4drive, yang2025worldrft}. Although more efficient, this approach sacrifices interpretable scene simulation and limits explicit visual verification that is valuable for safety-critical decision-making. Crucially, we identify a systemic limitation pervading both strategies: representation misalignment and task disconnection. Scene simulators are typically optimized for perceptual reconstruction, emphasizing vision representation, whereas planners are trained in isolation for action regression to encode motion representation. This lack of a unified representation --- one that is shared and consistent across both scene generation and planning tasks --- prevents the planner from fully leveraging the generative driving world model’s learned dynamics and motion priors.

\begin{figure}[t]
    \centering
    \includegraphics[width=1.0\linewidth]{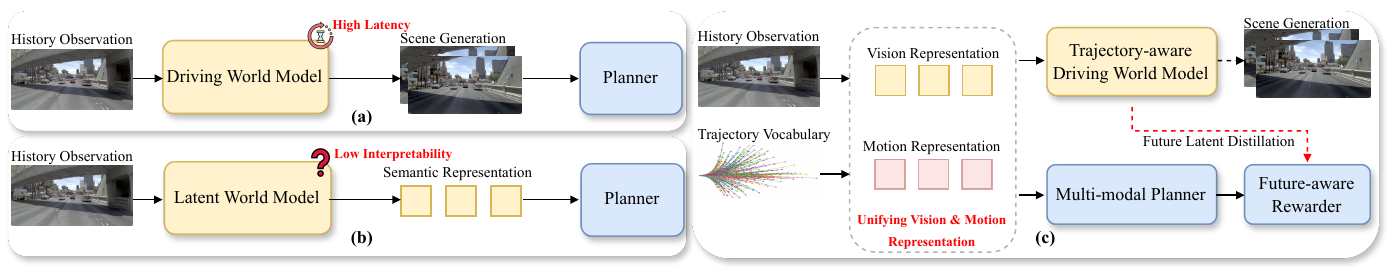}
    \caption{Driving world models for end-to-end autonomous driving. (a)~Planning with future scenes generated by a driving world model. (b)~Planning with semantic representation extracted from a latent world model. (c)~Proposed WorldDrive bridges planning and driving world model via unifying vision and motion representation.}
    \vspace{-6pt}
    \label{fig:intro}
\end{figure}

To bridge this gap, we introduce WorldDrive, a holistic framework designed to synergize end-to-end planning with scene generation via unifying vision and motion representation. The core philosophy of WorldDrive is representation unification: we posit that the representations learned through future generation~(scene generation) should be made directly reusable by the planner to decide the future~(planning). To this end, we first propose a Trajectory-aware Driving World Model~(TA-DWM). Unlike prior works that treat motion as a superficial condition, TA-DWM employs a multi-modal trajectory encoding scheme built on predefined trajectory anchors to construct a structured latent space where visual dynamics are intrinsically coupled with motion intentions. 

This design enables representation inheritance: the robust vision and motion encoders learned by the TA-DWM are directly transferred to initialize the downstream planner. This ensures that the planner operates in a mature and consistent latent space pre-aligned by the future scene generation task. Building upon this unified representation, we design a lightweight Multi-modal Planner. Leveraging these frozen encoders, the planner uses a query-centric cross-attention mechanism to efficiently fuse visual context with motion priors, generating diverse and high-quality trajectory candidates.

To harness the predictive foresight of the driving world model while avoiding the high latency of explicit video generation, we further introduce the Future-aware Rewarder~(FAR). Although TA-DWM is capable of synthesizing future videos, executing this generative process for every trajectory candidate is computationally prohibitive. Instead, FAR employs a planning-oriented distillation mechanism to directly distill the future latents from the frozen world model. This approach enables WorldDrive to evaluate candidate trajectories based on their corresponding distilled future latent. This design effectively aligns the trajectory selection with the world model’s learned dynamics while maintaining low-latency inference.

The main contributions of our work can be summarized as follows:
\begin{itemize}
    
\item We introduce a trajectory-aware driving world model that learns planner-shared vision and motion representations through trajectory-conditioned future generation. This design enables action-controllable future scene generation, producing plausible futures that are motion-consistent with the conditioning driving trajectory.
\item Leveraging the powerful representation learned by TA-DWM, we design a lightweight multi-modal planner that effectively fuses vision and motion cues. We also introduce a future-aware rewarder module that leverages the TA-DWM's foresight without the latency of explicit video generation, enabling real-time re-scoring of trajectory candidates at inference.
\item We integrate these components into WorldDrive, a holistic framework that bridges the representational schism between visual simulation and trajectory planning. Its design enables two capabilities: high-fidelity, motion-consistent future scene generation, and multi-modal, real-time planning.
\item We provide extensive evidence that WorldDrive achieves highly competitive world-model-based end-to-end planning performance while concurrently achieving high-fidelity performance on conditional scene generation tasks.
    
\end{itemize}

\section{Related Work}
\label{sec:related}

\subsection{End-to-end Autonomous Driving}
End-to-end autonomous driving aims to directly generate motion planning from raw sensor inputs and has attracted increasing research attention~\cite{review1, review2, hagedorn2024integration, hu2025vision}. Most end-to-end autonomous driving systems follow a general framework that integrates perception, prediction, and planning in a cascaded~\cite{uniad, fusionad} or parallel manner~\cite{paradrive, transfuser, fu2025orion} with a structured BEV feature~\cite{bevformer, li2024ego, diffusiondrive, gui2025trajdiff}. To reduce the reliance on dense BEV features, several works have proposed the use of sparse representations~\cite{vad, sparsead, sparsedrive, drivetransformer, wang2025omnidrive}. Leveraging driving priors, VADv2~\cite{vadv2} introduced a trajectory vocabulary as a prior for trajectory sampling. Building on this, Hydra-MDP~\cite{hydra} further distilled rule-based information. Additionally, DiffusionDrive~\cite{diffusiondrive} proposed a trajectory diffusion framework based on the trajectory vocabulary to accelerate the denoising process, and WoTE~\cite{li2025wote} utilized the trajectory anchor and future BEV state prediction to enhance the driving performance.

\subsection{Driving World Models}

Driving world models aim to predict the scene evolution from observations~\cite{review4, review5, review6, review7} and have shown great potential for generating long-horizon~\cite{magicdrivev2, vista, infinitydrive}, high-fidelity~\cite{drivedreamer, uniscene, drivingsphere, liang2025seeing, yan2025ad}, and corner-case scenario data~\cite{cosmos, terasim, sun2025terasim, zhou2025hermes}. Although these methods are capable of generating driving scenarios, driving world models need to simulate reasonable scenarios based on motion conditions~\cite{jia2023adriver, hu2023gaia, genad, hassan2025gem, li2025omninwm}. Vista~\cite{vista} achieved strong dynamic modeling by utilizing a larger volume of driving scenario data and introducing a motion loss. Building on this, DriVerse~\cite{driverse} realized trajectory-specific video generation by encoding trajectories as textual prompts and 2D spatial motion priors, and incorporated a motion alignment module. ReSim~\cite{yang2025resim} enriched real-world human demonstrations with diverse non-expert data collected from a driving simulator. DrivingGPT~\cite{drivinggpt} and Epona~\cite{epona} introduced a discrete action representation at each timestep on top of an autoregressive video generation framework, enabling controllable trajectory generation.

\subsection{World Model for Planning}
Research on optimizing planning policy using the world model has begun to show potential due to the strong representation capability~\cite{review4, review5, review8, li2025imagidrive}. Drive-WM~\cite{drivewm} used predicted trajectories as conditions for future scenario prediction and combined perception modules to evaluate safety. FSDrive~\cite{zeng2025fsdrive} proposed a visual chain-of-thought pipeline for future scene generation and planning. DrivingGPT~\cite{drivinggpt} and Epona~\cite{epona} used discrete and continuous autoregressive models to unify the generation of driving scenarios and driving trajectories, respectively. PWM~\cite{zhao2025pwm} and DriveVLA-W0~\cite{drivevlaw0} coupled trajectory planning with world simulation via action-free future state forecasting. Considering the high cost and latency of scene generation, latent world model methods, such as LAW~\cite{law}, World4Drive~\cite{world4drive}, and WorldRFT~\cite{yang2025worldrft}, demonstrated that effective scene representation can still be learned via self-supervision in latent space. In contrast, WorldDrive leverages world-model representations through feature inheritance and future-information distillation, enabling low-latency planning while retaining the benefits of predictive world-model representations.

\section{Method}
\begin{figure*}[t]
    \centering
    \includegraphics[width=1.0\linewidth]{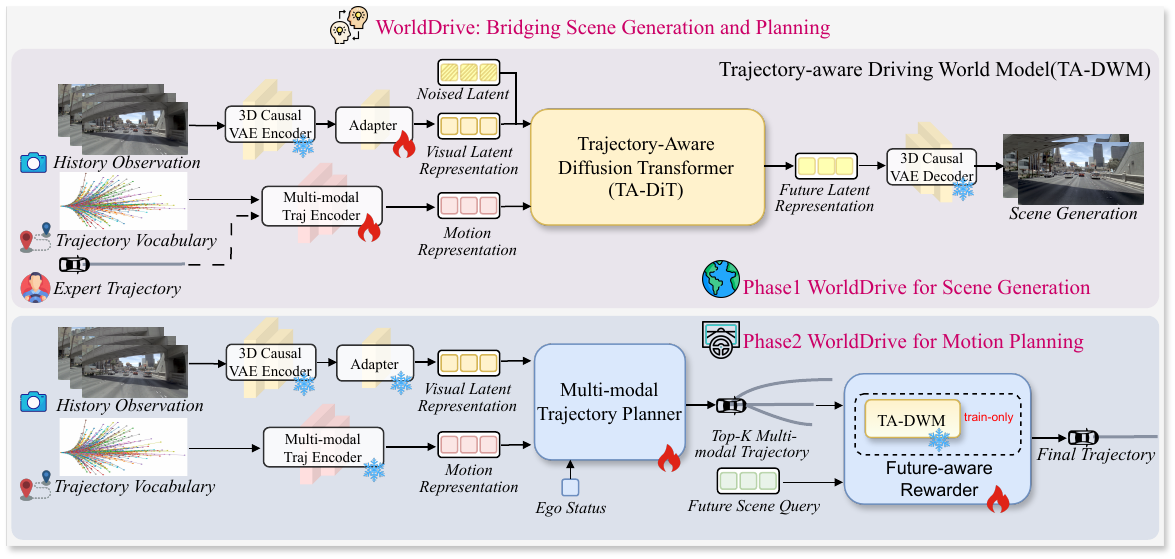}
    \caption{\textbf{Overall architecture of WorldDrive}. WorldDrive is a holistic framework unifying vision and motion representation to bridge scene generation and planning. The training process includes Phase 1: WorldDrive for scene generation, Phase 2: WorldDrive for motion planning. The vision and motion representations are optimized through the scene generation task. In the planning stage, the planner utilizes the frozen vision and trajectory encoders and outputs top-$K$ multi-modal trajectories. A future-aware rewarder is further designed to select the optimal trajectory from the candidates.}
    \vspace{-4pt}
    \label{fig:main}
\end{figure*}

As depicted in Fig.~\ref{fig:main}, WorldDrive bridges scene generation and planning by unifying vision and motion representation within a holistic framework. The system operates in two distinct yet coupled phases: Visual Simulation and Trajectory Optimization. In Phase 1, we develop a Trajectory-aware Driving World Model~(TA-DWM), employing a Trajectory-aware Diffusion Transformer~(TA-DiT) to synthesize future scenes. In Phase 2, the system transitions to planning. We inherit the frozen visual and trajectory encoders from Phase 1 to extract effective vision and motion representation. A learnable Multi-modal Trajectory Planner then generates diverse and high-quality trajectory candidates. A Future-aware Rewarder~(FAR) further identifies the optimal trajectory considering the predicted future scene latent representations.

\subsection{Trajectory-aware Driving World Model}
The trajectory-aware driving world model~(TA-DWM) builds upon a powerful video diffusion model, and we adapt it with a trajectory vocabulary as motion priors. As shown in Phase 1 of Fig.~\ref{fig:main}, we utilize a pre-trained 3D Causal VAE encoder to encode the historical observation~$x\in\mathbb{R}^{T \times 3 \times H \times W}$. To align the generic features with the driving domain, we introduce a lightweight and trainable visual adapter. The spatio-temporal visual latent $f$ is obtained as $f = \mathcal{E}_{\text{vis}}(x)$ where $ f\in \mathbb{R}^{C\times H' \times W'}$. To incorporate precise motion control, we propose a Multi-modal Trajectory Encoder that decomposes complex driving behaviors into coarse intentions and fine-grained adjustments. We first construct a trajectory vocabulary $\mathcal{V} \in \mathbb{R}^{N \times F \times 3}$ by clustering a large corpus of driving logs, where each primitive serves as a motion anchor. Given an expert trajectory $Y$, we retrieve the top-$M$ nearest anchors $\mathcal{V}_M \in \mathbb{R}^{M \times F \times 3} $ from the vocabulary and compute the residuals to capture motion details. The motion embedding $c$ is derived by fusing the anchor embeddings with residual embeddings:
\begin{equation}
    c = \mathcal{E}_{a}(\mathcal{V}_M) + \mathcal{E}_{o}(Y - \mathcal{V}_M),
    \label{equ:traj_embed}
\end{equation}
where $\mathcal{E}_{a}$ and $\mathcal{E}_{o}$ denote the anchor encoder and offset encoder, respectively. The resulting embedding $c \in \mathbb{R}^{M \times C}$ serves as the motion condition and is the multi-anchor decomposition of the input expert motion. TA-DiT is designed to generate future latents conditioned on historical context and motion intention. During training, the ground truth future frames $\bar{x}$ are encoded into the latent target $z_0 = \mathcal{E}_{\text{vis}}(\bar{x})$. We employ a standard forward diffusion process to obtain the noised state $z_t$. The TA-DiT,~$\epsilon_\theta({z_t; f, t, c})$, is then optimized to predict the added noise. The historical latent $f$ and motion embedding $c$ guide the generation to be both visually consistent and kinematically plausible. A detailed architectural illustration of the diffusion transformer is provided in the Appendix.

\subsection{Multi-modal Trajectory Planner}

Capitalizing on the structured latent space acquired in Phase 1, we initialize the planner by inheriting the frozen visual encoder and trajectory encoder from TA-DWM. This representation inheritance strategy ensures that the planner operates on effective and physically consistent features. To incorporate the vehicle state information, we employ a lightweight MLP to encode the ego status, including velocity, acceleration, and driving command into an embedding $e \in \mathbb{R}^{1 \times C}$. 

We formulate the trajectory planning as a set prediction problem.  We treat the complete predefined anchor embeddings $Q_{a} = \mathcal{E}_{a}(\mathcal{V}) \in \mathbb{R}^{N \times C}$ as queries. The context keys and values are constructed by concatenating the frozen visual latent $f$ and the ego embedding $e$. Through the attention mechanism, the planner aggregates relevant environmental context for each potential motion primitive:
\begin{equation}
    Q_{p} = \mathcal{D}_{\text{plan}}(Q_{a}, [f, e], [f, e]),
\end{equation}
where $[\cdot, \cdot]$ denotes concatenation along the sequence dimension, $\mathcal{D}_{\text{plan}}$ is the planning transformer decoder and $Q_{p}$ represents the enriched trajectory features. Adhering to the scoring paradigm of~\cite{li2025wote}, we employ an imitation reward model $\mathcal{D}_{\text{im}}$ and a simulation reward model $\mathcal{D}_{\text{sim}}$ to estimate the imitation reward and simulation reward of all trajectory anchors, and a regression network $\mathcal{D}_{\text{reg}}$ to predict the fine-grained offsets. 

Embracing the inherent multi-modality of driving scenarios, the planner outputs top-$K$ refined trajectories $\hat{Y}^{1:K} = \{\hat{Y}^{k}\}_{k=1}^{K} \in \mathbb{R}^{K \times F \times 3}$ by selecting the anchors with the highest top-$K$ combined scores and refining them via the predicted offsets. This strategy ensures the generation of a diverse trajectory distribution covering various plausible behaviors, thereby providing a comprehensive trajectory candidate set for the subsequent Future-aware Rewarder module.

\subsection{Driving with Future-aware Rewarder}

Although the planner already predicts scores for trajectories, these scores are computed mainly from the current scene representation. In principle, TA-DWM can provide predictive foresight by generating trajectory-conditioned future latents $\{z^k\}_{k=1}^K$ for candidates. However, running the full diffusion denoising process for all candidates is computationally prohibitive for real-time planning~\cite{drivewm, zeng2025fsdrive}. To retain the benefit of model-based foresight without performing expensive sampling at inference, we introduce the Future-aware Rewarder~(FAR). Rather than duplicating the planner score, FAR acts as a forecast-conditioned evaluator. During training, TA-DiT provides future latents conditioned on trajectory candidates $\hat{Y}^{1:K}$, and FAR is trained to distill the future-aware features. At inference, the planner first generates and ranks candidates, while FAR re-evaluates the top-$K$ candidates using the distilled future-aware cues, thereby bypassing the expensive diffusion rollout.

\begin{wrapfigure}{r}{0.53\linewidth}
  \vspace{-12pt}
  \centering
  \includegraphics[width=\linewidth]{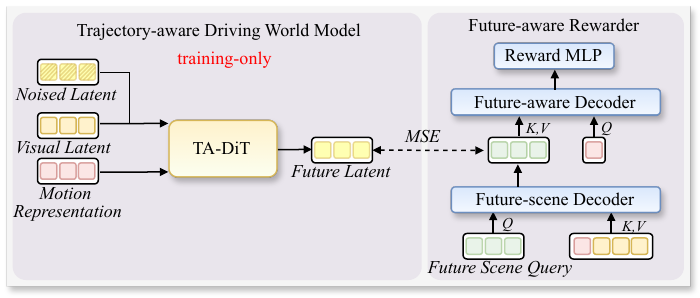}
  \caption{
    \textbf{Future-aware Rewarder~(FAR).} 
    During training, the frozen world model generates future latents. A distillation mechanism aligns the future scene queries with the future latents. During inference, the distilled features are directly queried by the motion representation to compute future-aware rewards.
  }
  \label{fig:rewarder}
\end{wrapfigure}

FAR introduces a set of learnable Future Scene Queries, $Q_{s} \in \mathbb{R}^{L \times C}$, as an information bottleneck for extracting planning-relevant future features. We employ a Future-scene Decoder that interacts with the shared historical visual latent $f$ and trajectory candidate embedding $c^k$ via cross-attention as:
\begin{equation}
    \hat{z}^k = \mathcal{D}_{\text{scene}}(Q_{s}, [f, c^k], [f, c^k]).
\end{equation}
During training, we align  $\hat{z}^k$ with the target future latents $ z^k $, and this design allows the lightweight decoder to effectively distill planning-relevant dynamics generated from the world model. A Future-aware Decoder then queries the distilled features using the trajectory embedding:
\begin{equation}
    h_{f}^k = \mathcal{D}_{\text{future}}(c^k, \hat{z}^k, \hat{z}^k),
\end{equation}
where $h_{f}^k$ is the future state feature for trajectory candidate.
Finally, an MLP maps $h_{f}^k$ to a scalar reward $r_k$, and we select the trajectory with the highest reward. By decoupling feature distillation from the heavy generation process, WorldDrive achieves real-time inference latency while ensuring that planning decisions are informed by future constraints.

\subsection{Training Loss}

We adopt a decoupled two-stage training curriculum to ensure the stable convergence of both the video diffusion transformer and the planning module. The first stage focuses exclusively on optimizing the TA-DWM to learn the latent dynamics conditioned on historical context $f$ and expert motion $c$. The objective follows the standard latent diffusion formulation:
\begin{equation}
    \mathcal{L}_{\text{world}} = \mathbb{E}_{z_0, t, \epsilon} \left[ \| {\epsilon} - \epsilon_\theta(z_t; f, t, c) \|^2_2 \right],
\end{equation}
where $\epsilon \sim \mathcal{N}(0, \mathbf{I})$ denotes the sampled Gaussian noise. 

In the second stage, we freeze the world model and optimize the multi-modal trajectory planner. We adhere to the supervision paradigm proposed in \cite{li2025wote}. The objective $\mathcal{L}_{\text{plan}}$ combines a cross-entropy imitation loss, a binary cross-entropy simulation loss, and an L1 regression loss for offset refinement of positive samples into a unified objective.

The training of FAR is governed by two complementary objectives: a feature distillation loss for latent alignment and a preference ranking loss for trajectory selection. First, to align the lightweight decoder with the world model's foresight, we minimize the L2 distance between the predicted features $\hat{z}^k$ and the target features $z^k$ generated by the frozen TA-DWM:
\begin{equation}
    \mathcal{L}_{\text{align}} = \mathbb{E}_{k} \left[ \| \hat{z}^k - \text{SG}(z^k) \|^2_2 \right],
\end{equation}
where $\text{SG}(\cdot)$ denotes the stop-gradient operator. For the scalar reward $r$, we model the trajectory selection as a preference ranking problem using the Bradley-Terry (BT) loss~\cite{btloss}. We construct preference pairs by contrasting the $\hat{Y}_{\text{pos}}$ against suboptimal candidates $\hat{Y}_{\text{neg}}$ in the top-$K$ candidates using an oracle driving score. The optimization objective is to maximize the likelihood that the positive sample achieves a higher reward score as:
\begin{equation}
    \mathcal{L}_{\text{reward}} = - \mathbb{E}_{(\hat{Y}_{\text{pos}}, \hat{Y}_{\text{neg}})\sim\hat{Y}^{1:K} } \left[ \log \sigma \left( r_{\text{pos}} - r_{\text{neg}} \right) \right],
\end{equation}  
where $\sigma(\cdot)$ is the sigmoid function. By minimizing $\mathcal{L}_{\text{reward}}$, FAR learns a scoring function that effectively discriminates between candidate trajectories. More details are provided in the Appendix.

\section{Experiments}

\subsection{Evaluation Settings}
For end-to-end planning task, we use NAVSIM~\cite{navsim}, NAVSIM-v2~\cite{cao2025pseudo}, nuScenes~\cite{nuscenes} to evaluate the performance. NAVSIM utilizes the Predictive Driver Model Score (PDMS) as the closed-loop metric. NAVSIM-v2 further introduces reactive traffic with pseudo closed-loop simulation, and extends PDMS with additional compliance metrics. These metrics provide a more comprehensive assessment of driving behavior. We also utilize nuScenes for a supplementary open-loop comparison to evaluate the performance. To assess the generation quality of the TA-DWM, we evaluate on the nuScenes validation split. Following common practices, we employ Fréchet Inception Distance (FID)~\cite{fid} and Fréchet Video Distance~\cite{fvd} to quantify the quality of generated scenes. Additionally, we conduct a quantitative analysis to evaluate the model's sensitivity to various motion controls. More closed-loop and ablation results are provided in the appendix.

\subsection{Implementation Details}
We initialize the 3D VAE and DiT from CogVideoX pretrained weights~\cite{cogvideox}. We construct the trajectory vocabulary using K-means clustering and set the number of trajectory anchors to 256, following WoTE~\cite{li2025wote}. In Phase 1, TA-DWM is trained on a combined dataset of driving videos from nuPlan~\cite{caesar2021nuplan} and nuScenes~\cite{nuscenes}. Optimization is performed on 16 NVIDIA A100 GPUs with a batch size of 32. In Phase 2, we freeze the encoders and train the planner for 50 epochs on the NAVSIM navtrain split using 8 NVIDIA 3090 GPUs with a batch size of 256. To train FAR, we freeze the trajectory planner and optimize FAR for 10 epochs using PDMS as the oracle for preference supervision. At inference time, WorldDrive does not perform explicit future scene generation, enabling low-latency planning. Additional implementation details are provided in the supplementary material.

\begin{table}[t]
        \centering
        \footnotesize
        \caption{\textbf{NAVSIM navtest split comparison.}~PDMS and sub-scores reflecting closed-loop performance. MV: multi-view cameras; SV: single-view camera; L: LiDAR. $\dagger$: Training with full navtrain split. $\ddagger$: Best-of-6 performance with oracle.}
        \begin{tabular}{l|c|c|cccccc}
        \toprule
        Method & Sensors & WM-Based & NC & DAC & TTC & Comf. & EP & PDMS \\
        \midrule
        TransFuser~\cite{transfuser} & MV \& L & $\times$ & 97.7 & 92.8 & 92.8 & 100.0 & 79.2 & 84.0 \\
        Hydra-MDP~\cite{hydra} & MV \& L & $\times$ & 98.3 & 96.0 & 94.6 & 100.0 & 78.7 & 86.5 \\
        DiffusionDrive~\cite{diffusiondrive} & MV \& L & $\times$ & 98.2 & 96.2 & 94.7 & 100.0 & 82.2 & 88.1 \\
        WoTE~\cite{li2025wote} & MV \& L & $\checkmark$ & 98.5 & 96.8 & 94.9 & 99.9 & 81.9 & 88.3 \\
        \midrule
        LAW~\cite{law} & MV & $\checkmark$ & 96.4 & 95.4 & 88.7 & 99.9 & 81.7 & 84.6 \\
        World4Drive~\cite{world4drive} & MV & $\checkmark$& 97.4 & 94.3 & 92.8 & \textbf{100.0} & 79.9 & 85.1 \\
        FSDrive~\cite{zeng2025fsdrive} & MV & $\checkmark$& 98.2 & 93.8 & 93.3 & 99.9 & 80.1 & 85.1 \\
        DrivingGPT~\cite{drivinggpt} & SV & $\checkmark$ & \textbf{98.9} & 90.7 & 94.9 & 95.6 & 79.7 & 82.4 \\
        Epona~\cite{epona} & SV & $\checkmark$ & 97.9 & 95.1 & 93.8 & 99.9 & 80.4 & 86.2 \\
        ImagiDrive~\cite{li2025imagidrive} & SV & $\checkmark$ & 98.6 & \textbf{96.2} & 94.5 & \textbf{100.0} & 80.5 & 87.4 \\
        \textbf{WorldDrive} & SV & $\checkmark$ & 98.4 & \textbf{96.2} & \textbf{95.1}  & \textbf{100.0} & \textbf{81.9} & \textbf{88.1} \\
        \midrule
        PWM~\cite{zhao2025pwm}$\dagger$ & SV & $\checkmark$ & 98.6 & 95.9 & 95.4 & 100.0 & 81.8 & 88.1 \\
        DriveVLA-W0~\cite{drivevlaw0}$\dagger$ & SV & $\checkmark$ & \textbf{98.7} & 96.2 & \textbf{95.5} & 100.0 & 82.2 & 88.4 \\

        \textbf{WorldDrive}$\dagger$ & SV & $\checkmark$ & 98.4 & \textbf{97.0} & 95.1  & \textbf{100.0} & \textbf{83.4} & \textbf{89.2} \\
        \midrule
        \textcolor{gray}{DriveVLA-W0$\ddagger$} 
        & \textcolor{gray}{SV} 
        & \textcolor{gray}{$\checkmark$} 
        & \textcolor{gray}{\textbf{99.3}} 
        & \textcolor{gray}{97.4} 
        & \textcolor{gray}{97.0} 
        & \textcolor{gray}{\textbf{100.0}} 
        & \textcolor{gray}{\textbf{88.3}} 
        & \textcolor{gray}{93.0} \\
        
        \textcolor{gray}{\textbf{WorldDrive}$\ddagger$} 
        & \textcolor{gray}{SV} 
        & \textcolor{gray}{$\checkmark$} 
        & \textcolor{gray}{\textbf{99.3}} 
        & \textcolor{gray}{\textbf{98.8}} 
        & \textcolor{gray}{\textbf{97.9}} 
        & \textcolor{gray}{\textbf{100.0}} 
        & \textcolor{gray}{\textbf{88.3}} 
        & \textcolor{gray}{\textbf{93.6}} \\
        \bottomrule
        \end{tabular}
\vspace{-1mm}
\label{tab:closeloop_turn}
\end{table}

\begin{table*}[t!]
\centering
\footnotesize
\caption{\textbf{NAVSIM-v2 comparison.} EPDMS and sub-scores reflecting closed-loop performance on both the navtest and navhard splits.}
\vspace{-2mm}
\label{tab:navsimv2_comparison}

\textbf{(a) NAVSIM-v2 navtest split}
\vspace{1mm}

\begin{tabular}{l | c c c c c c c c c | c}
    \toprule
    Method 
    & NC
    & DAC
    & DDC
    & TLC
    & EP
    & TTC
    & LK
    & HC
    & EC
    & EPDMS \\
    \midrule
    TransFuser~\cite{transfuser} & 96.9 & 89.9 & 97.8 & 99.7 & 87.1 & 95.4 & 92.7 & 98.3 & 87.2 & 76.7 \\
    DiffusionDrive~\cite{diffusiondrive} & 98.2 & 95.9 & 99.4 & 99.8 & 87.5 & 97.3 & 96.8 & 98.3 & 87.7 & 84.5 \\
    \midrule
    DriveVLA-W0~\cite{drivevlaw0} & \textbf{98.5} & \textbf{99.1} & 98.0 & 99.7 & 86.4 & \textbf{98.1} & 93.2 & 97.9 & 58.9 & 86.1 \\
    \textbf{WorldDrive} & \textbf{98.5} & 96.7 & \textbf{99.0} & \textbf{99.8} & \textbf{87.2} & \textbf{98.1} & \textbf{95.8} & \textbf{98.3} & \textbf{79.2} & \textbf{87.6} \\
    \bottomrule
\end{tabular}

\vspace{2mm}

\textbf{(b) NAVSIM-v2 navhard split}
\vspace{1mm}

\resizebox{\textwidth}{!}{
\begin{tabular}{l  | c | c c c c c c c c c | c}
    \toprule
    Method 
    & Stage
    & NC
    & DAC
    & DDC
    & TLC
    & EP
    & TTC
    & LK
    & HC
    & EC
    & EPDMS \\
    \midrule
    LTF~\cite{transfuser}  & \makecell{S1 \\ S2} &
    \makecell{96.2 \\ 77.7} & \makecell{79.5 \\ 70.2} & \makecell{\textbf{99.1} \\ 84.2} & \makecell{99.5 \\ 98.0} & \makecell{84.1 \\ 85.1} & \makecell{95.1 \\ 75.6} & \makecell{94.2 \\ 45.4} & \makecell{97.5 \\ 95.7} & \makecell{79.1 \\ \textbf{75.9}} & 23.1 \\
    \midrule
    DiffusionDrive~\cite{diffusiondrive}  & \makecell{S1 \\ S2} &
    \makecell{96.8 \\ 80.1} & \makecell{86.0 \\ 72.8} & \makecell{98.8 \\ 84.4} & \makecell{99.3 \\ 98.4} & \makecell{\textbf{84.0} \\ \textbf{85.9}} & \makecell{95.8 \\ 76.6} & \makecell{\textbf{96.7} \\ 46.4} & \makecell{\textbf{97.6} \\ \textbf{96.3}} & \makecell{\textbf{79.6} \\ 72.8} & 27.5 \\
    \midrule
    \textbf{WorldDrive}  & \makecell{S1 \\ S2} &
    \makecell{\textbf{97.3} \\ \textbf{91.4}} & \makecell{\textbf{89.1} \\ \textbf{82.0}} & \makecell{97.6 \\ \textbf{91.0}} & \makecell{\textbf{99.7} \\ \textbf{98.5}} & \makecell{60.5 \\ 53.1} & \makecell{\textbf{96.8} \\ \textbf{90.6}} & \makecell{87.7 \\ \textbf{52.3}} & \makecell{93.1 \\ 93.3} & \makecell{60.0 \\ 62.8} & \textbf{34.9} \\
    \bottomrule
\end{tabular}
}

\vspace{-2mm}
\end{table*}


\begin{table*}[!t]
\centering
\begin{minipage}[t]{0.52\textwidth}
\centering
\footnotesize
\setlength{\tabcolsep}{5pt}
\caption{\textbf{Planner with different pretrained representation}. ``VAE'' means initializing with the pretrained 3D VAE. ``Vis.'' and ``Mot.'' mean initializing the vision and motion encoders from TA-DWM.}
\label{tab:rep}
\vspace{1mm}
\begin{tabular}{ccc|cccc|c}
\toprule
VAE & Vis. & Mot. & NC & DAC & TTC & EP  & PDMS  \\ 
\midrule
$\times$ & $\times$ & $\times$ & 69.0 & 57.6 & 57.2 & 28.7  & 31.4 \\
$\checkmark$ & $\times$ & $\times$ & 97.6 & 93.6 & 93.2 & 79.5  & 84.9 \\
$\checkmark$ & $\checkmark$ & $\times$ & 97.9 & 94.1 & 93.3 & \textbf{80.7}  & 85.8 \\
$\checkmark$ & $\checkmark$ & $\checkmark$ & \textbf{98.4} & \textbf{95.1} & \textbf{94.7} & 80.4 & \textbf{86.9} \\
\bottomrule
\end{tabular}
\vspace{-6mm}
\end{minipage}
\hfill
\begin{minipage}[t]{0.47\textwidth}
\centering
\footnotesize
\caption{\textbf{Future-aware rewarder design}. ``Traj. feat.'' means using candidate trajectory features from the planner. ``Future feat.'' means using future scene queries and future latent distillation.}
\label{tab:rewarder}
\setlength{\tabcolsep}{4.5pt}
\vspace{1mm}
\begin{tabular}{cc|cccc|c}
\toprule
Traj. & Future & \multirow{2}{*}{NC} & \multirow{2}{*}{DAC} & \multirow{2}{*}{TTC} & \multirow{2}{*}{EP} & \multirow{2}{*}{PDMS}  \\ 
feat. & feat.  &  &  &  &  &  \\
\midrule
$\times$ & $\times$ & 98.4 & 95.1 & 94.7 & 80.4 & 86.9 \\
$\checkmark$ & $\times$ & 98.3 & 95.3 & 94.3 & 81.2 & 87.0 \\
$\checkmark$ & $\checkmark$ & \textbf{98.4} & \textbf{96.2} & \textbf{95.1} & \textbf{81.9} & \textbf{88.1} \\
\bottomrule
\end{tabular}
\vspace{-6mm}
\end{minipage}
\end{table*}

\begin{figure*}[t]
  \begin{minipage}[t]{0.45\textwidth}
  \centering
  \vspace{0pt}
  \footnotesize
  \setlength{\tabcolsep}{3pt}
    \captionof{table}{\textbf{nuScenes validation split open-loop planning comparison}. We follow the SparseDrive\cite{sparsedrive} evaluation metric. V: multi-view camera; SV: single-view camera.}
    \renewcommand{\arraystretch}{0.95}
    \begin{tabular}{l|c|cc|cc}
    \toprule
    \multirow{2}{*}{Method} &
    \multirow{2}{*}{Sensors} & 
    \multicolumn{2}{c|}{L2~($m$) $\downarrow$} & 
    \multicolumn{2}{c}{CR~(\%) $\downarrow$} \\
     &   & 3$s$ & Avg.  & 3$s$ & Avg.\\
    \midrule
    UniAD~\cite{uniad} & V  & 1.04  & 0.73  & 0.63 & 0.61 \\
    VAD~\cite{vad} & V   & 1.05 & 0.72   & 0.43 & 0.21\\
    SparseDrive~\cite{sparsedrive} & V   & 0.96 & 0.61  & 0.18 & 0.08\\
    \midrule
    LAW~\cite{law} & V   & 1.01 & 0.61   & 0.54 & 0.30 \\
    World4Drive~\cite{world4drive} & V   & 0.81 & 0.50   & \textbf{0.33} & 0.16 \\
    Drive-WM~\cite{drivewm} & V   &  1.20 & 0.80   & 0.48 & 0.26 \\
    Epona~\cite{epona} & SV   & 1.98 & 1.25   & 0.85 & 0.36 \\
    \textbf{WorldDrive} & SV  & \textbf{0.68} & \textbf{0.42}  & 0.38 & \textbf{0.16} \\
    \bottomrule
    \end{tabular}
    \vspace{-3mm}
    \label{tab:openloop_turn}
    \end{minipage}
    \hfill
    \begin{minipage}[t]{0.51\textwidth}
    \vspace{0pt}
    \centering
    \footnotesize
    \includegraphics[width=0.95\textwidth]{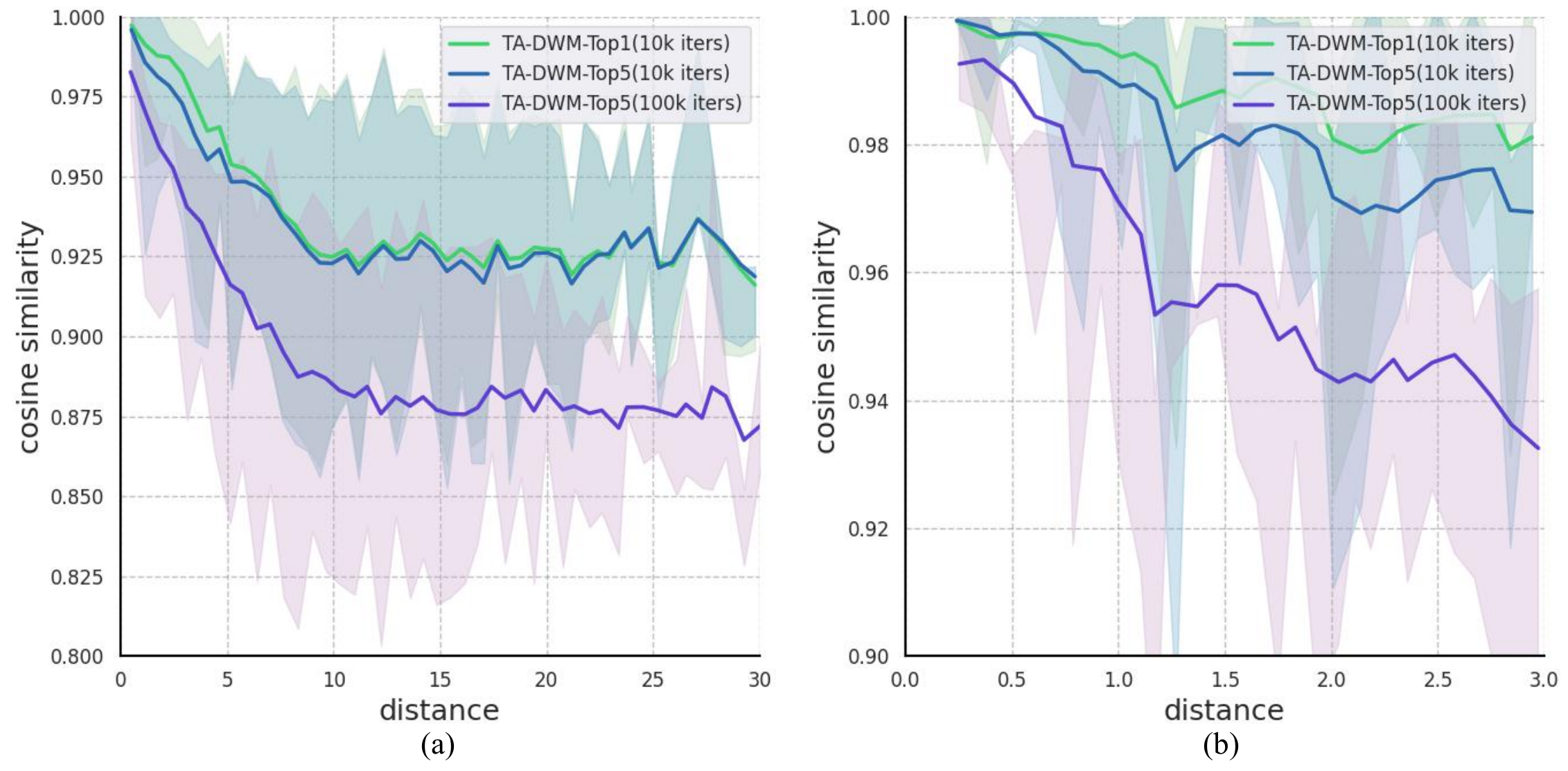}
    \caption{\textbf{Quantitative Analysis of Motion Sensitivity.}~The similarity between scene representations is inversely correlated with the geometric distance. The sensitivity to both (a)~large and (b)~small deviations is amplified with further training.}
    \label{fig:similarity}
    \vspace{-3mm}
  \end{minipage}
\end{figure*}

\renewcommand{\arraystretch}{0.95}
\begin{table}[t]
    \centering
    \footnotesize
    \caption{\textbf{Comparison of prediction fidelity on nuScenes validation split generated videos.}}
    \resizebox{1.0\textwidth}{!}{
    \begin{tabular}{c|cccccc |c}
    \toprule
    Metric       & DriveDreamer~\cite{drivedreamer} & WoVoGen~\cite{wovogen} & GenAD~\cite{genad}   & Drive-WM~\cite{drivewm} & Vista~\cite{vista} & Driverse~\cite{driverse} & WorldDrive \\
    \midrule
    Resolution     &  128$\times$192 &  256$\times$448 &  256$\times$448 &  192$\times$384 &  480$\times$832 &  480$\times$832 & 256$\times$512 \\
    \midrule
    FID            &     52.6     &   27.6   &   15.4   &  15.2  &   18.2   &   20.1   & 12.8  \\
    FVD            &     452.0    &   417.7  &   184.0  &  122.7 &   158.0  &   143.5  & 131.7 \\
    \bottomrule
    \end{tabular}}
    \label{fig:gen}
    \vspace{-6mm}
\end{table}

\subsection{WorldDrive on Trajectory Planning}
\textbf{Performance comparison with SOTA methods.}~Table~\ref{tab:closeloop_turn} compares WorldDrive with state-of-the-art methods on the NAVSIM navtest split. WorldDrive surpasses the leading single-view method ImagiDrive~\cite{li2025imagidrive}, and leading multi-view approaches such as FSDrive~\cite{zeng2025fsdrive}. WorldDrive also outperforms the strong world model methods PWM~\cite{zhao2025pwm} and DriveVLA-W0~\cite{drivevlaw0}. In the best-of-N setting, we use the oracle scorer to select the best trajectory and WorldDrive achieves 93.6 PDMS. This oracle result is reported to probe the upper bound and to verify that WorldDrive can generate a multi-modal candidate set, and it is not used for inference. Table~\ref{tab:navsimv2_comparison} reports results on NAVSIM-v2 navtest and navhard split. The gains are consistent across most safety-related metrics. We observe that some comfort metrics remain challenging, suggesting further effort for improving long-horizon efficiency while preserving safety. Table~\ref{tab:openloop_turn} reports the open-loop comparison on nuScenes. Among world-model-based methods, WorldDrive achieves highly competitive performance. These results corroborate the effectiveness of WorldDrive, confirming that the synergistic optimization of visual and motion representations within the TA-DWM translates into robust planning capabilities.

\textbf{Impact of Pre-training Strategy.} Table~\ref{tab:rep} studies the effect of pretrained representations in WorldDrive. The pretrained 3D Causal VAE provides strong visual priors and substantially improves PDMS. Inheriting the TA-DWM vision and motion representations further yields step-wise improvements, validating the effectiveness of representation inheritance for downstream planning.

\textbf{Trajectory Rewarder Strategy Analysis.}~A central challenge in multi-modal planning is selecting the best trajectory from diverse candidates. Table~\ref{tab:rewarder} shows that using trajectory features alone brings only marginal gain and even slightly degrades some safety metrics. FAR improves PDMS from 86.9 to 88.1 by distilling future latents from TA-DWM. This demonstrates that FAR provides complementary future-aware ranking signals rather than redundant re-scoring.

\renewcommand{\arraystretch}{0.95}
\begin{wraptable}{r}{0.45\textwidth}
\vspace{-12pt}
\centering
\footnotesize
\caption{\textbf{Inference latency comparison}. Latency is measured on a single NVIDIA A800 GPU. ``PWM $\dagger$'' means PWM without future frame forecast. ``WorldDrive $\dagger$'' means WorldDrive without future-aware rewarder.}
\vspace{-2pt}
\setlength{\tabcolsep}{2pt}
\begin{tabular}{c|c c c c c|c}
\toprule
Method & NC & DAC & TTC & EP & PDMS & Latency\\
\midrule
PWM$\dagger$ & 98.0 & 95.1 & 94.1 & 82.4 & 87.3 & 570ms \\
WorldDrive$\dagger$ & 98.3 & 95.8 & 95.1  & 81.2 & 87.8 & 36ms \\ 
PWM & 98.6 & 95.9 & 95.4 & 81.8 & 88.1 & 850ms \\
WorldDrive & 98.4 & 97.0 & 95.1  & 83.4 & 89.2 & 53ms \\ 
\bottomrule
\end{tabular}
\vspace{-25pt}
\label{tab:latency}
\end{wraptable}
\textbf{Inference Latency Analysis.}~Table~\ref{tab:latency} reports inference efficiency. DWM-based planners face a practical trade-off: incorporating future forecasts improves performance but increases inference latency. WorldDrive avoids this by distilling future latents and eliminating explicit future scene generation at inference.
\vspace{-2mm}

\subsection{WorldDrive on Scene Generation}
\textbf{Generation Quality on nuScenes.}~Table~\ref{fig:gen} presents quantitative results on nuScenes. WorldDrive achieves competitive generation quality with structured trajectory conditioning, suggesting that a robust pretrained VAE combined with structured motion priors can yield high-fidelity simulations.


\textbf{Motion sensitivity.}~To quantify TA-DWM controllability, we measure the cosine similarity between generated latents conditioned on different trajectories and correlate it with their geometric distance. Fig.~\ref{fig:similarity} shows a clear inverse correlation where latent similarity decreases as motion deviation increases, and this sensitivity is amplified with longer training from 10k to 100k iterations. We also show the impact of the multi-modal trajectory encoder. Top-5 multi-modal trajectory encoder conditioning further improves discrimination over the Top-1 baseline, especially in the small-deviation regime.
\vspace{-0.5mm}

\begin{figure*}[t]
    \centering
    \includegraphics[width=0.95\linewidth]{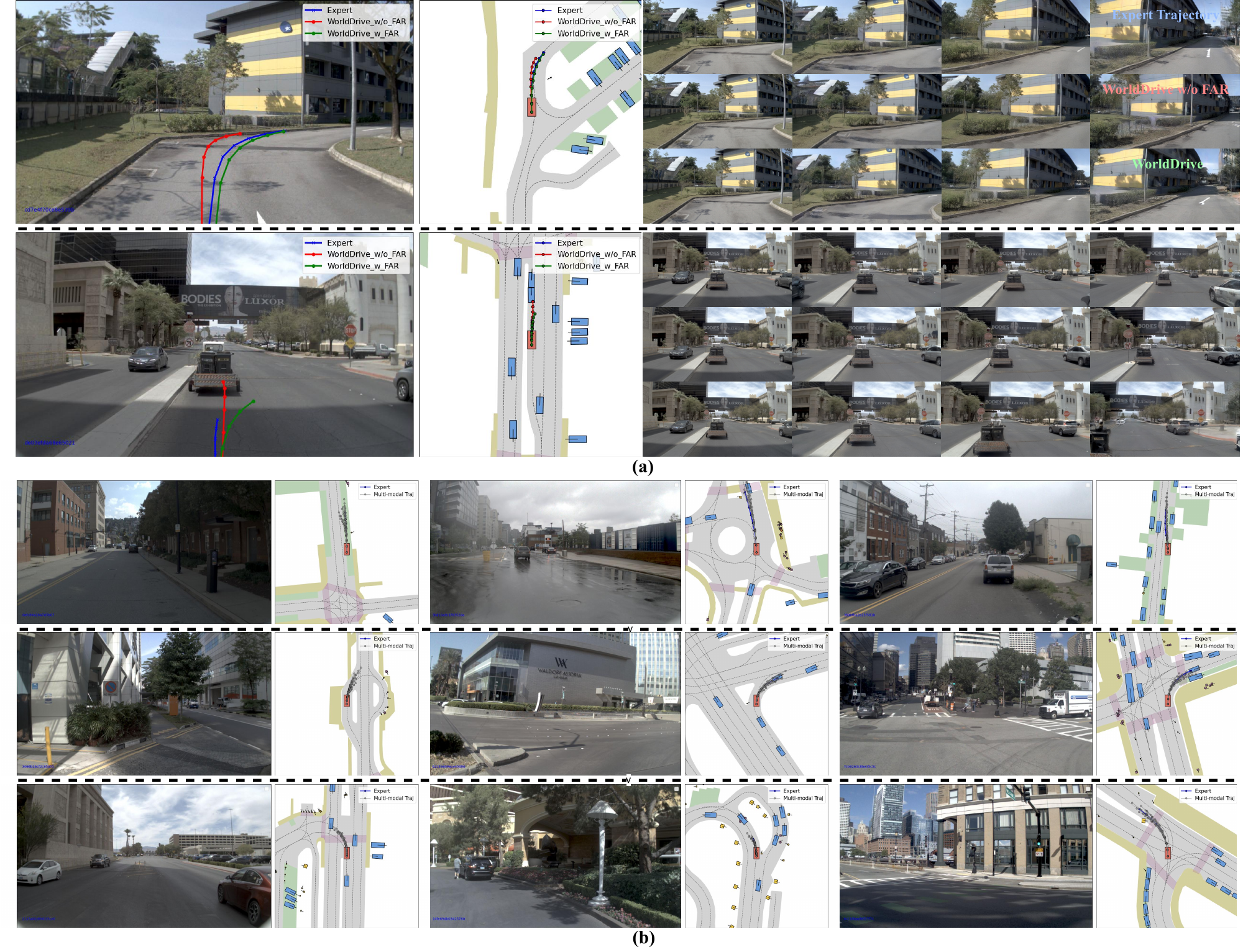}
    \vspace{-2mm}
    \caption{\textbf{Qualitative planning result on NAVSIM navtest split}. \textbf{(a)}~Planning result and the corresponding generated future scene with different trajectories. \textbf{(b)}~Top-10 Multi-modal trajectories.}
    \label{fig:planning}
    \vspace{-2mm}
\end{figure*}

\begin{figure*}[tb]
    \centering
    \includegraphics[width=0.95\linewidth]{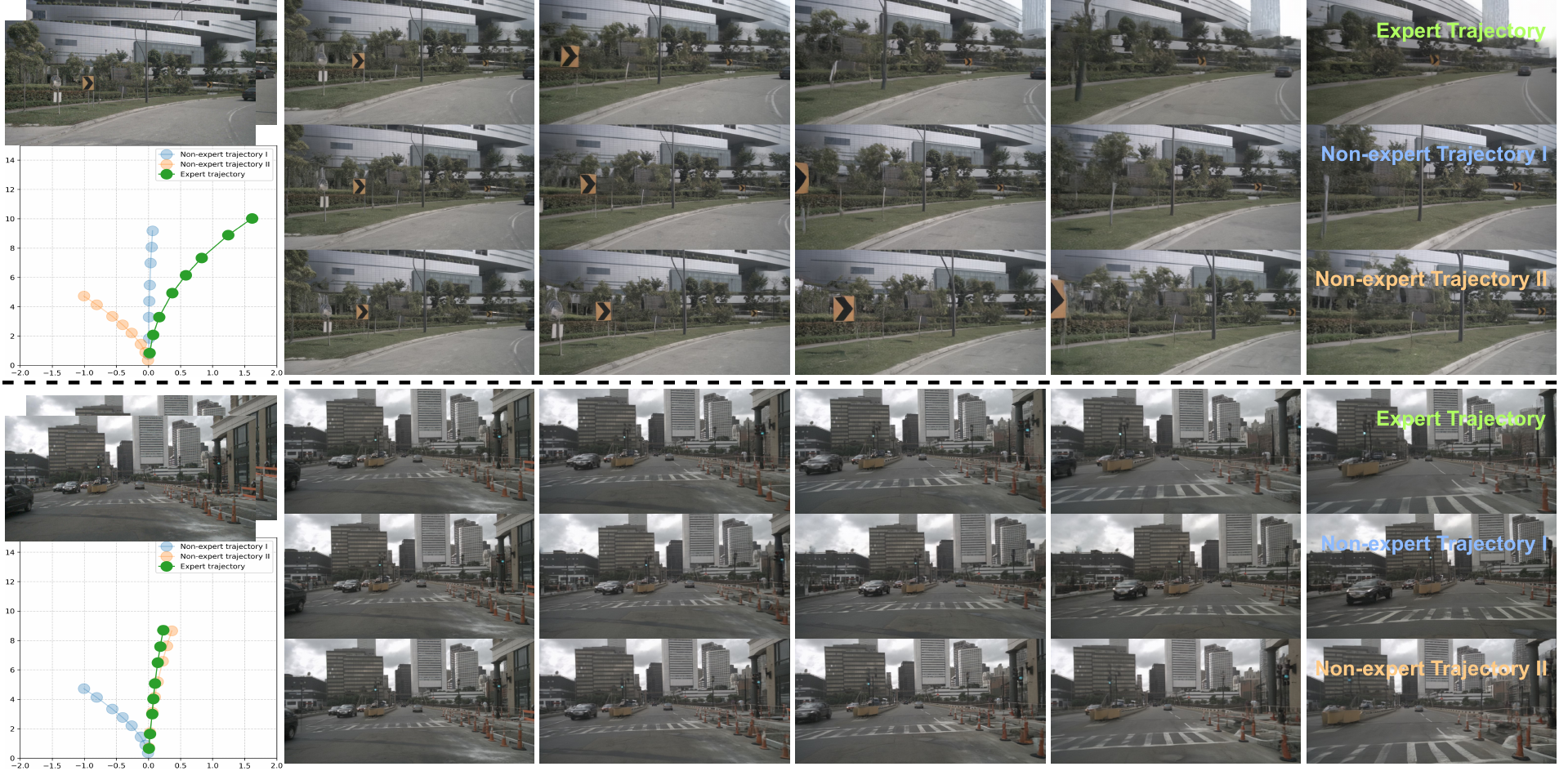}
    \caption{\textbf{Qualitative simulation result on nuScenes}. Scenes are generated with three trajectory conditions: the expert trajectory (green), and non-expert alternatives.}
    \label{fig:visual}
\vspace{-1mm}
\end{figure*}


\subsection{Qualitative Results}
\vspace{-0.5mm}

Fig.~\ref{fig:planning} visualizes the qualitative planning results, where WorldDrive produces safe and feasible trajectories and TA-DWM can be invoked for future-scene case analysis. Fig.~\ref{fig:planning}(b) further shows the predicted multi-modal trajectories. Fig.~\ref{fig:visual} presents the generated scene results under different motion conditions, showing that the generated futures closely follow the specified motion and remain plausible even for challenging trajectories toward non-drivable areas or key traffic participants.



\section{Conclusion}
We present WorldDrive, a unified framework that bridges the representational gap between scene generation and end-to-end planning. By introducing the Trajectory-aware Driving World Model, we establish a feature space where visual dynamics are coupled with motion intentions. Building on this unified representation, we introduce representation inheritance that initializes a lightweight planner with mature features learned from scene generation. We further propose the Future-aware Rewarder, which leverages DWM foresight while maintaining real-time inference. Extensive experiments show that WorldDrive achieves strong planning performance and supports action-controllable scene synthesis. We hope this work inspires future research on generative world models as a foundation for safe and interpretable autonomous driving.

\section{Acknowledgments and Disclosure of Funding}

This work was supported by the Science and Technology Development Fund of Macau SAR (FDCT) under grants 0134/2025/RIA2 and 0102/2023/RIA2, and the Jiangyin Hi-tech Industrial Development Zone under the Taihu Innovation Scheme (EF2025-00003-SKL-IOTSC)


\bibliographystyle{plain}
\bibliography{ref}
\newpage

\clearpage
\appendix
\vskip8pt
\noindent\textbf{\textit{\LARGE Appendix}}
\vskip8pt

The Appendix is organized as follows:
\begin{itemize}
    \item Sec.~\ref{sec:app_wm_comparison}: positioning of WorldDrive among world-model-based planners.
    \item Sec.~\ref{sec:supp_implementation}: implementation details for model architectures and training.
    \item Sec.~\ref{sec:supp_results}: additional closed-loop planning results and ablation studies.
    \item Sec.~\ref{sec:supp_qualitative}: additional qualitative results.
    \item Sec.~\ref{sec:supp_limit}: limitations and future directions.
\end{itemize}

\section{Comparison with World-Model-based Planning Methods}
\label{sec:app_wm_comparison}

To further clarify the positioning of WorldDrive, we compare it with representative world-model-based planning methods in Table~\ref{tab:wm_comparison}. Existing methods typically exploit world models as future generators, simulators, or latent representation learners. In contrast, WorldDrive makes world-model representations directly usable by the planner through representation inheritance, and further distills future-aware information into a lightweight rewarder to avoid expensive online rollout.

\begin{table}[h]
\centering
\footnotesize
\caption{\textbf{Comparison with recent world-model-based planners.} 
``Future scene gen.'' denotes whether the method supports explicit  future scene generation. 
``Traj.-cond.'' denotes whether the world model is explicitly conditioned on trajectory candidates. 
``Rep. inheritance'' denotes whether world-model representations are directly reused by the downstream planner. 
``No online rollout'' denotes whether the method avoids explicit pixel-level future generation or rollout during inference.}
\label{tab:wm_comparison}
\setlength{\tabcolsep}{3.5pt}
\renewcommand{\arraystretch}{1.05}
\begin{tabular}{lcccc}
\toprule
Method & Future scene gen. & Traj.-cond. & Rep. inheritance & No pixel-level online rollout \\
\midrule
Drive-WM~\cite{drivewm}      & $\checkmark$ & $\checkmark$ & $\times$ & $\times$ \\
FSDrive~\cite{zeng2025fsdrive} & $\checkmark$ & $\checkmark$ & $\times$ & $\times$ \\
LAW~\cite{law}              & $\times$     & $\times$     & $\times$ & $\checkmark$ \\
World4Drive~\cite{world4drive} & $\times$ & $\checkmark$ & $\times$ & $\checkmark$ \\
PWM~\cite{zhao2025pwm}      & $\checkmark$ & $\times$ & $\times$ & $\times$ \\
DriveVLA-W0~\cite{drivevlaw0} & $\checkmark$ & $\times$ & $\times$ & $\checkmark$ \\
DrivingGPT~\cite{drivinggpt} & $\checkmark$ & $\checkmark$ & $\times$ & $\times$ \\
Epona~\cite{epona}          & $\checkmark$ & $\checkmark$ & $\checkmark$ & $\times$ \\
\midrule
WorldDrive                  & $\checkmark$ & $\checkmark$ & $\checkmark$ & $\checkmark$ \\
\bottomrule
\end{tabular}
\end{table}

\section{Further Implementation Details}
\label{sec:supp_implementation}
\subsection{Trajectory-aware Driving World Model}
\begin{figure}[t]
    \centering
    \includegraphics[width=1.0\linewidth]{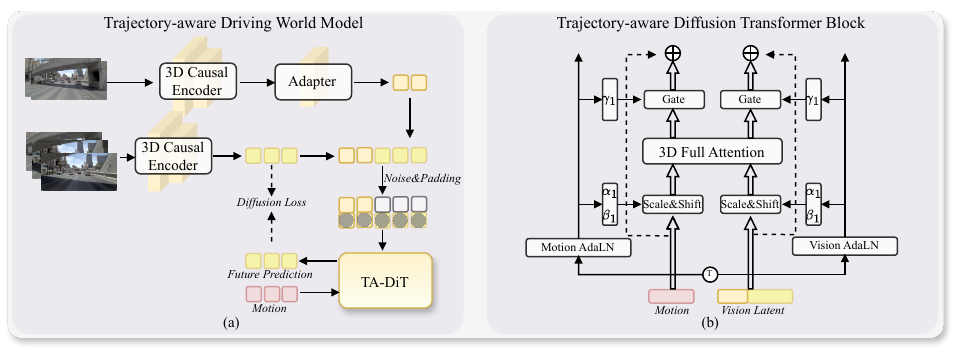}
    \caption{The details of (a)~Trajectory-aware Driving World Model and (b)~Trajectory-aware Diffusion Transformer block.}
    \label{fig:sup1}
\end{figure}

A comprehensive overview of the architecture and data flow of the trajectory-aware driving world model is illustrated in Figure~\ref{fig:sup1}(a). The pipeline first encodes both the historical context frames and the future target frames into a compact latent space using a pre-trained and frozen 3D Causal VAE. The historical latent features are further processed by a learnable Visual Adapter, which refines spatial-temporal representations for effective conditioning. The target for the diffusion process, $z_0$, is constructed by concatenating the latent representations of both the historical and future frames along the temporal axis. Following the standard diffusion framework, we inject Gaussian noise to obtain $z_t$. The Trajectory-Aware Diffusion Transformer (TA-DiT) is then optimized to predict the noise, conditioned on the adapted historical context, the diffusion timestep $t$, and the multi-modal trajectory embedding. Specifically, the motion embedding $c$ is injected into the transformer blocks, functioning analogously to text prompts in the original CogVideoX framework to provide high-level guidance for the generation process. The internal structure of the TA-DiT block is illustrated in Figure~\ref{fig:sup1}(b).

\subsection{Multi-modal Trajectory Planner}
Adhering to the protocol proposed in~\cite{li2025wote}, we adopt two types of supervision for each trajectory anchor. The imitation reward supervision $r_{\text{im}}$ measures the distance between the trajectory anchors and the expert trajectory, while the simulation reward supervision $r_{\text{sim}}$ assesses the trajectory quality based on safety and efficiency rules as:
\begin{equation}
    r_{\text{sim}} = \{r_{\text{sim}}^{\text{NC}},r_{\text{sim}}^{\text{DAC}}, r_{\text{sim}}^{\text{TTC}}, r_{\text{sim}}^{\text{Comf}}, r_{\text{sim}}^{\text{EP}} \},
\end{equation}
where the sub-metrics represent no collisions (NC), drivable area compliance (DAC), time-to-collision (TTC), comfort (Comf), and ego progress (EP). The final reward is defined as the weighted log-sum of these rewards as:
\begin{equation}
\begin{split}
    r_{\text{plan}} = - \big( & \omega_1 \text{log}r_{\text{im}} + \omega_2 \text{log}r_{\text{sim}}^{\text{NC}} + \omega_3\text{log} r_{\text{sim}}^{\text{DAC}} +\\ &
    \omega_4 \text{log}(5r_{\text{sim}}^{\text{TTC}} 
     +  2r_{\text{sim}}^{\text{Comf}} +  5r_{\text{sim}}^{\text{EP}}) \big),
\end{split}
\label{eq:reward_plan}
\end{equation}
where $\omega_i$ are the hyper-parameters and are set as $\omega_1 = 0.1$, $\omega_2 = \omega_3 = 0.5$, $\omega_4 = 1$. For the supervision, the target of the imitation reward is calculated based on the \text{L2} distance $d_i$ between the trajectory anchor and expert trajectory and the softmax function as $r_\text{im}^* = \frac{\text{exp}(-d_i)}{\sum_{j=1}^N\text{exp}(-d_j)}$. The predicted imitation scores are supervised using the Cross-Entropy (CE) loss against this target as:
\begin{equation}
    \mathcal{L}^{\text{im}} = \text{CrossEntropy}(r_{im}, r^*_{im}).
\end{equation} 
Regarding the simulation rewards, we use the simulator to produce five rewards for evaluating a trajectory and use the Binary Cross-Entropy loss to supervise the predicted simulation rewards as:
\begin{equation}
    \mathcal{L}^{\text{sim}} = \text{BCE}(r_{sim}, r^*_{sim}).
\end{equation}
For the offset regression, we identify the positive anchor, which is the closest to the expert and supervise its predicted offset using an \text{L1} regression loss.

\subsection{Future-aware Rewarder}
\label{sup:FAR}
A critical aspect of training the FAR is the construction of effective preference pairs. To strike a balance between training efficiency and hard-negative mining, we adopt a sampling strategy. For each scenario, we first run the planner in inference mode to generate the top-16 candidate trajectories. From this candidate set, we select the top-1 trajectory, which is the one with the highest oracle score. Three trajectories with the lowest oracle scores are selected as hard negatives, and three randomly sampled trajectories from the remaining pool are selected to increase sample diversity.

These selected trajectories form preference pairs for optimization via the Bradley--Terry (BT) model loss. During the inference phase, the trajectory candidate with the highest reward score is chosen as the final output. We keep FAR as a separate module rather than directly integrating it into the planner for two reasons. First, the planner and FAR play different roles: the planner generates feasible trajectory proposals from the current observation, while FAR estimates their future consequences using distilled world-model foresight. Second, applying future-aware evaluation only to a small top-$K$ set avoids requiring future prediction for all motion anchors, which preserves low-latency inference. This decomposition allows WorldDrive to exploit the predictive capability of TA-DWM while maintaining a lightweight planning pipeline.

\subsection{Training Details}
\noindent\textbf{Driving World Model Pre-training.}~The driving world model pre-training is based on CogVideoX-2B. TA-DWM is pre-trained in two sequential stages. In the first stage, the model is trained for 200,000 iterations on the nuPlan training dataset at a $256 \times 512$ resolution. The objective is to predict 17 future frames from a context of 8 historical frames, with all video clips sampled at 10Hz. In the second stage, we fine-tune the model on the nuScenes dataset for an additional 100k iterations under the same resolution and frame rate settings. A constant learning rate of $1 \times 10^{-4}$ is employed throughout both stages. All generation metrics reported in the main paper are evaluated using this nuScenes-adapted model. The details are shown in Table~\ref{tab:tadwm_training_setting}.

\begin{table*}[h]
\centering
\footnotesize
\caption{\textbf{Training data and settings of TA-DWM.}~The details include the data usage, temporal configuration, and optimization settings.}
\label{tab:tadwm_training_setting}
\setlength{\tabcolsep}{2.5pt}
\renewcommand{\arraystretch}{1.05}
\begin{tabular}{l|c|c|c|c|c|c}
\toprule
Stage & Dataset & Iter. & Resolution & Input / Target & Frame Rate & LR \\
\midrule
\multirow{2}{*}{TA-DWM pre-training}
& nuPlan 
& 200k 
& $256 \times 512$ 
& 8 hist. frames $\rightarrow$ 17 future frames 
& 10Hz 
& $1{\times}10^{-4}$ \\

& nuScenes 
& 100k 
& $256 \times 512$ 
& 8 hist. frames $\rightarrow$ 17 future frames 
& 10Hz 
& $1{\times}10^{-4}$ \\
\midrule
Representation fine-tuning 
& NAVSIM 
& 100k 
& $512 \times 1024$ 
& 4 hist. frames $\rightarrow$ 9 future frames 
& 2Hz 
& $5{\times}10^{-5}$ \\
\bottomrule
\end{tabular}
\end{table*}

\noindent\textbf{Planner and Rewarder Training.}~The training process for the planner and rewarder module involves three distinct steps. The process begins with a representation fine-tuning stage, where the pre-trained TA-DWM, including vision and motion encoders, is further trained on the NAVSIM dataset for 100k iterations at a higher resolution of $512 \times 1024$ with a learning rate of $5 \times 10^{-5}$, following the setting in~\cite{epona}. The model takes 4 historical frames (2s) to predict a 4.5s horizon (9 frames), with padding applied to meet the 3D Causal VAE's spatial constraints. Initialized with the adapted encoder weights, which are kept frozen, the Multi-modal Planner is trained for 50 epochs on the NAVSIM navtrain training set using a cosine learning rate scheduler with a peak learning rate of $6 \times 10^{-4}$. Finally, the Future-aware Rewarder (FAR) is trained for 10 epochs with a learning rate of $3 \times 10^{-4}$. The training samples are constructed using the sampling strategy detailed in Sec.~\ref{sup:FAR}.

\section{Further Experiments}
\label{sec:supp_results}
\subsection{Closed-loop Planning and Simulation Results with CARLA Simulator}

\begin{figure*}[t]
    \centering
    \includegraphics[width=1.0\linewidth]{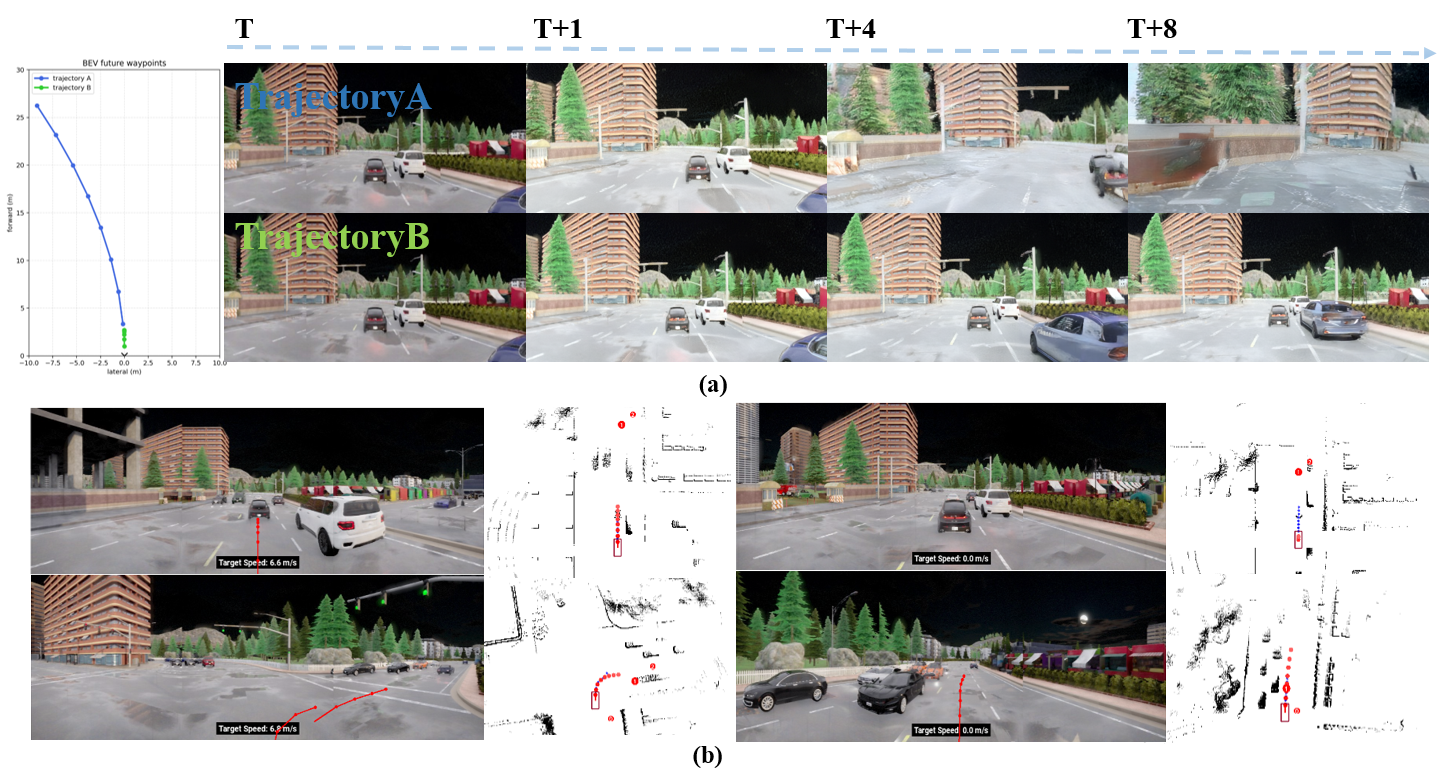}
    \caption{Qualitative results of (a)zero-shot action-controllable future scene generation and (b)end-to-end planning result on Bench2Drive with WorldDrive}
    \label{fig:bench2drive}
\end{figure*}

Since closed-loop planning evaluation still relies on simulators rather than real-world driving scenarios, we additionally evaluate WorldDrive on the Bench2Drive\cite{jia2024bench2drive} benchmark, which is built upon CARLA simulator\cite{dosovitskiy2017carla}. We directly reuse the NAVSIM-adapted TA-DWM and incrementally validate the transferability of the pre-trained visual latent representations. We instantiate the Bench2Drive planner using the DriveTransformer~\cite{drivetransformer} training protocol, while augmenting the visual representation with the NAVSIM-adapted TA-DWM visual encoder. The planner is trained with Think2Drive\cite{li2024think2drive} expert supervision. The results are reported in Table~\ref{tab:close-loop} and in Fig.\ref{fig:bench2drive}(b). We also provide a future scene generation result of WorldDrive in the CARLA environment in Fig.\ref{fig:bench2drive}(a), which can be seen as a zero-shot simulation. When there are counterfactual trajectories, the generated scene can still capture the corresponding driving intentions and trajectories.

\begin{table*}[h]
\footnotesize
\caption{\textbf{Bench2Drive closed-loop planning comparison}. DS: Driving Score, SR: Success Rate. WorldDrive uses Think2Drive~\cite{li2024think2drive} as the expert.}
\label{tab:close-loop}
\setlength{\tabcolsep}{5pt}
\centering
\begin{tabular}{lcccc}
\toprule
\multirow{2}{*}{Method}  & \multicolumn{4}{c}{Closed-loop Metric} \\
 & DS & SR(\%) & Efficiency & Comfortness \\
\midrule
AD-MLP~\citep{zhai2023rethinking}  & 18.05 & 0.00 & 48.45 & 22.63  \\
UniAD-Tiny~\citep{uniad}  & 40.73 & 13.18 & 123.92 & 47.04 \\
UniAD-Base~\citep{uniad}  & 45.81 & 16.36 & 129.21 & 43.58 \\
VAD~\citep{vad}  & 42.35 & 15.00 & 46.01 & 0.91 \\
MomAD~\citep{song2025don} & 44.54 & 16.71 & 170.21 & 48.63 \\
GenAD~\citep{genad}  & 44.81 & 15.90 & - & - \\
DriveTransformer-Large~\citep{drivetransformer} & 63.46 & 35.01 & 100.64 & 20.78 \\
\midrule
\textbf{WorldDrive} & 69.76 & 37.27 & 92.26 & 36.67 \\
\bottomrule
\end{tabular}
\end{table*}

\subsection{Further Ablation Studies}

\begin{table}[t]
    \centering
    \footnotesize
    \begin{minipage}[t]{0.49\textwidth}
    \caption{\textbf{Ablation on trajectory candidates.}
    Top-K indicates the trajectories with the K highest predicted scores from the multi-modal planner.}
    \label{tab:topk_anchor}
        \centering
        \vspace{1mm}

        \begin{tabular}{c|c c c c|c}
        \toprule
        Top-K & NC & DAC & TTC & EP & PDMS \\
        \midrule
        1  & 98.4 & 95.1 & 94.7 & 80.4 & 86.9 \\
        3  & 98.4 & 96.0 & 95.2 & 81.5 & 87.9 \\
        5  & 98.4 & 96.2 & 95.1 & 81.9 & 88.1 \\
        10 & 98.0 & 95.9 & 94.4 & 81.9 & 87.6 \\
        \bottomrule
        \end{tabular}
    \end{minipage}
    \hfill
    \begin{minipage}[t]{0.48\textwidth}
        \centering
        \caption{\textbf{Ablation on trajectory vocabulary size.}
    Vocab denotes the number of trajectories in the trajectory vocabulary.}
    \label{tab:vocab}
        \vspace{1mm}

        \begin{tabular}{c|c c c c|c}
        \toprule
        Vocab & NC & DAC & TTC & EP & PDMS \\
        \midrule
        64  & 98.3 & 95.2 & 94.7 & 80.2 & 86.8 \\
        128 & 98.4 & 96.1 & 95.0 & 81.6 & 87.9 \\
        256 & 98.4 & 96.2 & 95.1 & 81.9 & 88.1 \\
        512 & 98.4 & 96.1 & 95.1 & 82.0 & 88.1 \\
        \bottomrule
        \end{tabular}
    \end{minipage}
\end{table}

\noindent\textbf{Top-K Candidates in FAR.}~Table~\ref{tab:topk_anchor} investigates the sensitivity of the Future-aware Rewarder to the number of input trajectory candidates. Increasing K from 1 to 5 yields consistent improvements in PDMS, suggesting that a larger candidate set covers more diverse driving modes and enables the rewarder to select more efficient and compliant trajectories.
However, further increasing K to 10 leads to a performance drop. This indicates that an excessive number of trajectory candidates may introduce more low-quality trajectories, which can distract the rewarder and compromise safety metrics. Therefore, we set K=5 as the default choice, which provides the best trade-off between diversity and precision.

\noindent\textbf{Trajectory Vocabulary.}~Table~\ref{tab:vocab} presents the ablation on the size of the trajectory vocabulary. Overall, increasing the vocabulary size consistently improves planning performance from 64 to 256 entries, with PDMS rising from 86.8 to 88.1. A small vocabulary is likely insufficient to capture the diversity of motion intentions. When further increasing the vocabulary size to 512, the performance becomes saturated. We use 256 trajectory anchors as the default setting.

\noindent\textbf{WorldDrive training stability.}
To verify the reproducibility of WorldDrive, we repeat the experiments multiple times and report the mean and standard deviation in Table~\ref{tab:stability}. 
The WorldDrive baseline without FAR already exhibits stable performance, with small standard deviations across metrics, showing that the multi-modal planner based on inherited world-model representations is reliable. 
Moreover, FAR brings consistent improvements while maintaining low variance, increasing PDMS from 86.89 to 88.14. 
The gains on DAC, TTC, and EP indicate that FAR contributes complementary future-aware evaluation beyond the planner's original scores, verifying that FAR is non-redundant and effectively refines trajectory selection.

\noindent\textbf{Alternative FAR target.}
To examine whether FAR's gain depends on PDMS-based preference supervision, we replace the oracle with trajectory ADE and the rule-based ego-progress submetric. Table~\ref{tab:far_oracle}
shows that both alternatives improve PDMS over the baseline.

\begin{table}[t]
\centering
\footnotesize
\caption{\textbf{WorldDrive training stability comparison over five runs.} 
The standard deviation is highlighted in blue.}
\begin{tabular}{l|c c c c|c}
\toprule
Method & NC & DAC & TTC & EP & PDMS \\
\midrule
WorldDrive (w/o FAR) 
& 98.35{\textcolor{blue}{$\pm$0.12}} 
& 95.16{\textcolor{blue}{$\pm$0.15}} 
& 94.63{\textcolor{blue}{$\pm$0.22}} 
& 80.56{\textcolor{blue}{$\pm$0.28}} 
& 86.89{\textcolor{blue}{$\pm$0.15}} \\

WorldDrive 
& 98.36{\textcolor{blue}{$\pm$0.08}} 
& 96.14{\textcolor{blue}{$\pm$0.14}} 
& 95.03{\textcolor{blue}{$\pm$0.27}} 
& 82.06{\textcolor{blue}{$\pm$0.21}} 
& 88.14{\textcolor{blue}{$\pm$0.09}} \\
\bottomrule
\end{tabular}
\label{tab:stability}
\end{table}

\begin{table}[!htbp]
\centering
\footnotesize
\caption{\textbf{FAR target comparison}. Ablation with alternative preference oracles.}
\label{tab:far_oracle}
\setlength{\tabcolsep}{4pt}
\begin{tabular}{l | ccc | c}
\toprule
Oracle & NC & DAC & EP & PDMS \\
\midrule
Baseline & 98.4 & 95.1 & 80.4 & 86.9 \\
ADE      & 98.5 & 96.0 & 80.7 & 87.5 \\
EP-only  & 98.4 & 95.5 & 82.0 & 87.8 \\
\bottomrule
\end{tabular}
\end{table}

\begin{table}[!htbp]
\centering
\footnotesize
\caption{\textbf{Inference latency comparison}. Latency is measured on a single NVIDIA A800 GPU for the full forward pipeline of WorldDrive. ``Encoders'' include the 3D Causal VAE, visual adapter, and trajectory encoder.}
\setlength{\tabcolsep}{4pt}
    \begin{tabular}{c|c c c |c}
    \toprule
    Module & Encoders & Planner & Rewarder & Total\\

    \midrule
    Latency & 17.9ms & 18.9ms & 16.4ms & 53.2ms \\
    \bottomrule
\end{tabular}
\label{sup:latency}
\end{table}

\begin{figure*}[!htbp]
    \centering
    \includegraphics[width=1.0\linewidth]{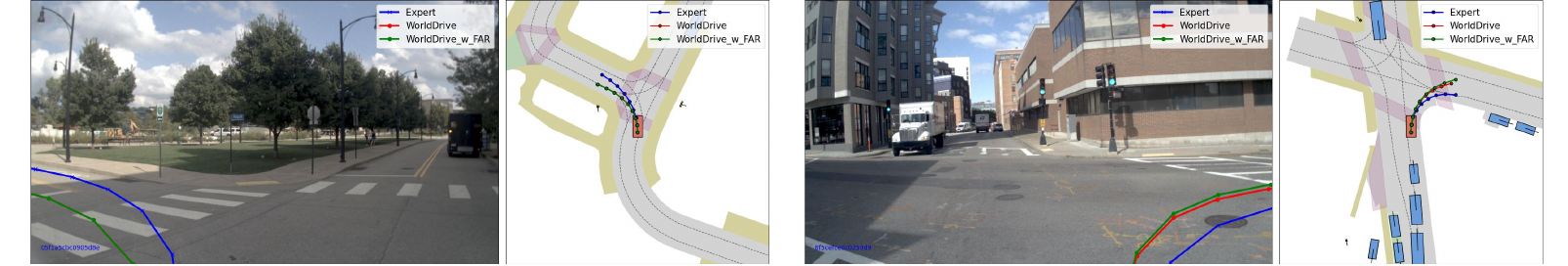}
    \caption{Qualitative results of the failure case of WorldDrive.}
    \label{sup:failure}
\end{figure*}

\noindent\textbf{Feasibility Analysis.}~We present the detailed latency of each part of WorldDrive in Table~\ref{sup:latency}. We set the batch size to 1 and the trajectory candidate number used in FAR to 5. Both the multi-modal planner and the FAR module are designed for architectural efficiency, comprised primarily of lightweight transformer decoders and MLP layers. This design facilitates exceptionally low-latency execution.  The results confirm that when operating in the planning inference mode, WorldDrive meets low-latency requirements.

\section{Further Qualitative Result}
\label{sec:supp_qualitative}

\noindent\textbf{Failure Cases.}~We provide a representative failure case of WorldDrive in Figure~\ref{sup:failure}. 
We observe that WorldDrive may fail to maintain proper lane keeping in turning scenarios with blind spots or occlusions. 
This limitation is partly due to the single-view input setting and the lack of multi-perspective or multi-modal observations during TA-DWM training. 
Such failure cases suggest that incorporating multi-view cameras, multi-sensor inputs, or stronger occlusion-aware world modeling could be a promising direction for future research.

\begin{figure*}[t]
    \centering
    \includegraphics[width=1.0\linewidth]{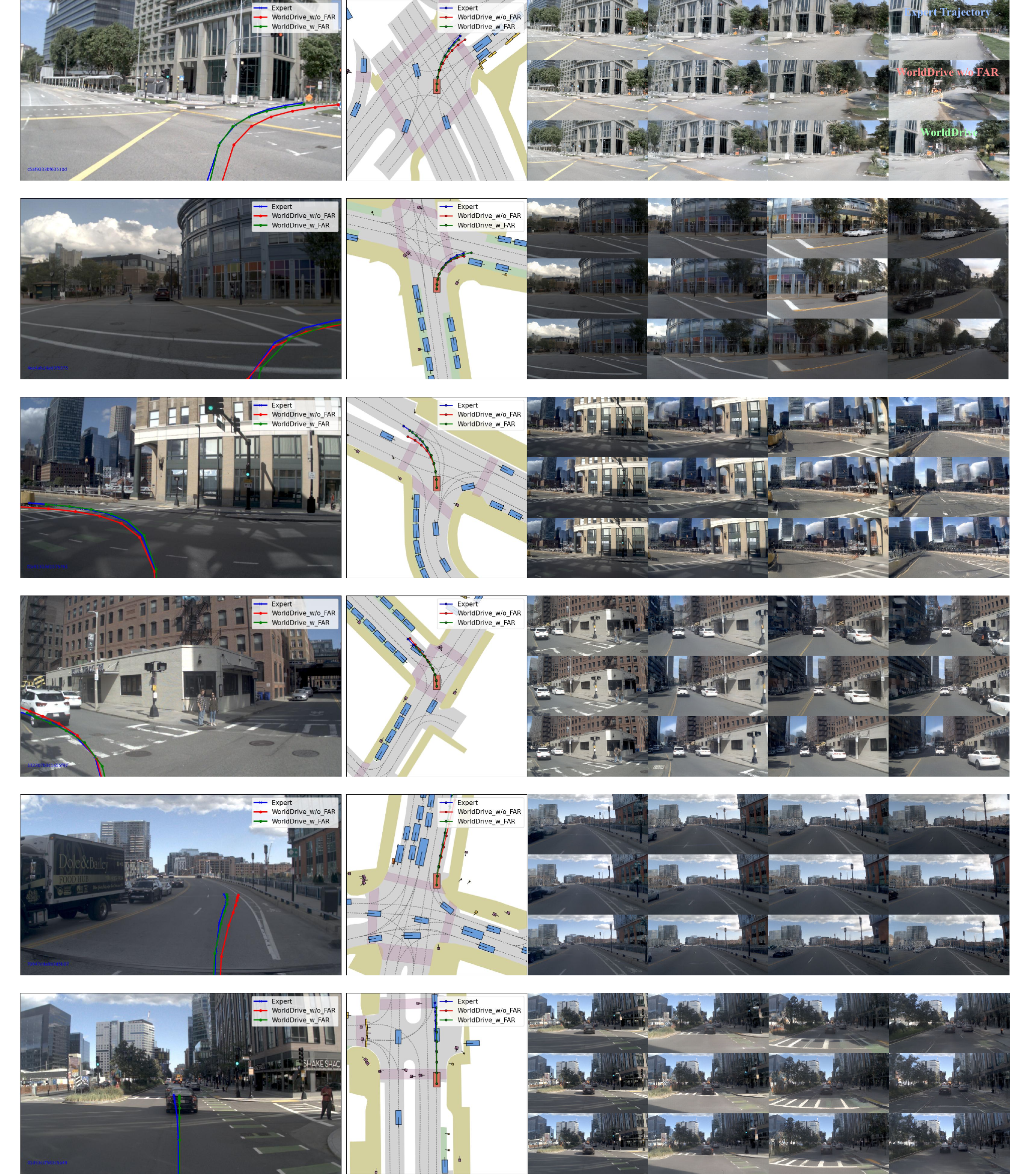}
    \caption{Qualitative results of planning and future scene generation with WorldDrive on the NAVSIM \textit{navtest} split.}
    \label{fig:sup2}
\end{figure*}

\noindent\textbf{Visualization of Planning and Scene Generation.}~Fig.~\ref{fig:sup2} provides additional qualitative results demonstrating the planning and generation capabilities of WorldDrive on the NAVSIM dataset. Benefiting from the unified representation, the planner initially proposes a set of high-quality trajectory candidates that closely align with the expert's decision. However, deviations may occur in complex scenarios. FAR effectively mitigates these risks by identifying and selecting the most favorable behavior, which demonstrates better alignment with the expert trajectory. In addition, we feed the selected trajectory into TA-DiT and present the generated future scene sequence. The results show that TA-DiT can synthesize realistic future videos that remain consistent with the planned trajectory.

\begin{figure*}[t]
    \centering
    \includegraphics[width=1.0\linewidth]{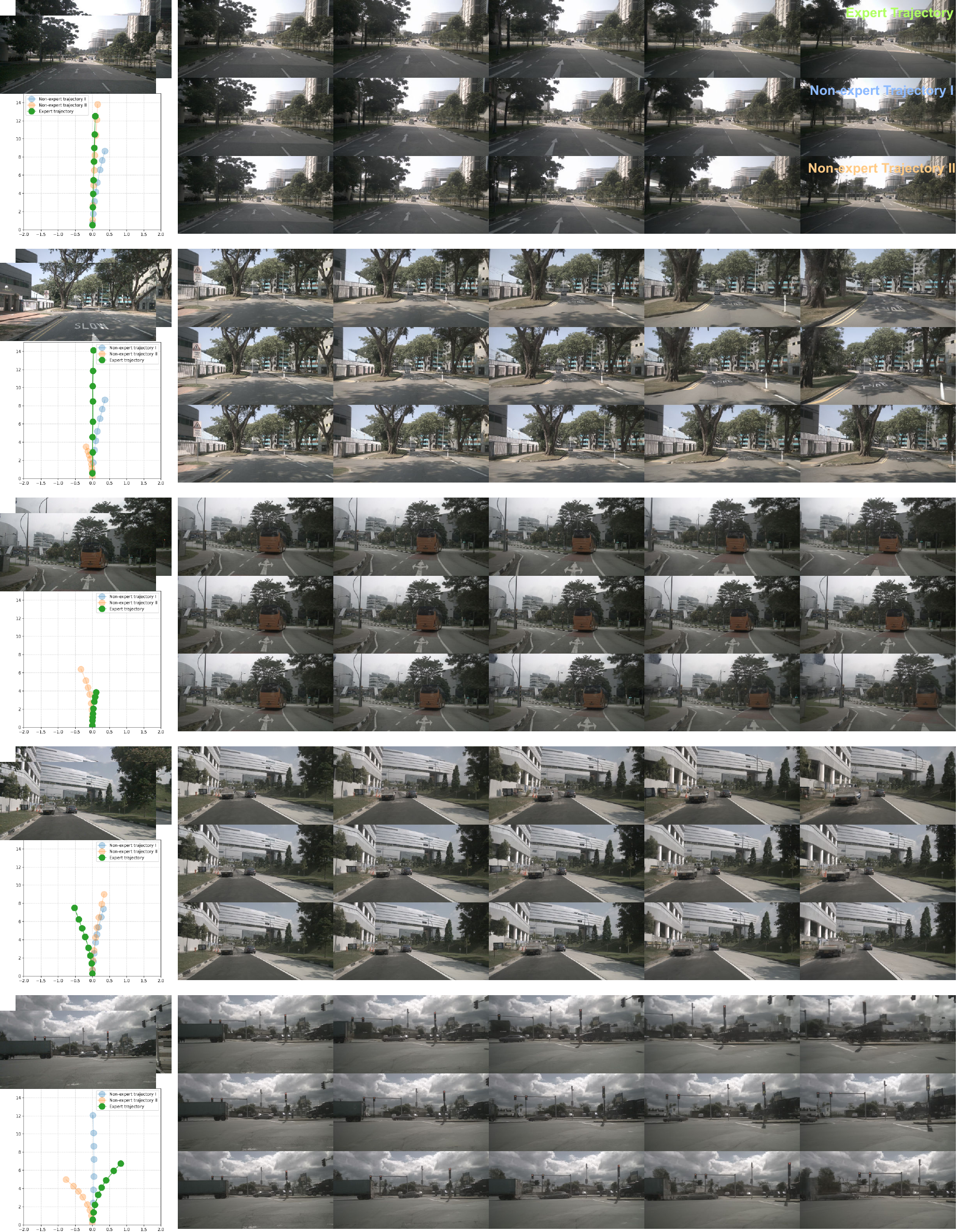}
    \caption{Qualitative results of action-controllable future scene generation with WorldDrive on the nuScenes validation set.}
    \label{fig:sup3}
\end{figure*}

\noindent\textbf{Visualization of Action-Controllable Scene Simulation.}~Fig.~\ref{fig:sup3} demonstrates the controllability of Trajectory-aware Driving World Model (TA-DWM). Specifically, we feed the different trajectories into TA-DWM to synthesize the corresponding expected future outcomes. As observed in the generated sequences, the synthesized scenes exhibit remarkable geometric consistency with the input motion commands. This confirms that TA-DWM does not merely memorize video textures but captures motion-consistent scene dynamics, enabling it to support future-scene case analysis for diverse planning decisions.

\section{Limitations}
\label{sec:supp_limit}
Despite its strong performance, WorldDrive still has several limitations. First, it only uses front-view camera observations, without multi-view or multi-sensor inputs, which is a weaker sensor configuration than conventional end-to-end driving methods using multi-view cameras or multi-sensor fusion. This may limit its robustness in scenarios requiring wide-range spatial perception, such as large intersections and side-agent interactions. 
Second, although the trajectory vocabulary provides structured and effective motion priors, it is constructed from the training trajectory distribution and may be less expressive for rare maneuvers, highly interactive behaviors, or extreme out-of-distribution scenarios. 

In future work, we will explore world-model-based planners with multi-view or multi-sensor inputs to further enhance performance and investigate the integration with other forms of action representation.

\newpage

\clearpage
\section*{NeurIPS Paper Checklist}



\begin{enumerate}

\item {\bf Claims}
    \item[] Question: Do the main claims made in the abstract and introduction accurately reflect the paper's contributions and scope?
    \item[] Answer:\answerYes{} 
    \item[] Justification: The abstract and introduction clearly outline the contribution and proposed methodological frameworks.

\item {\bf Limitations}
    \item[] Question: Does the paper discuss the limitations of the work performed by the authors?
    \item[] Answer: \answerYes{} 
    \item[] Justification: We discuss in the Appendix. 

\item {\bf Theory assumptions and proofs}
    \item[] Question: For each theoretical result, does the paper provide the full set of assumptions and a complete (and correct) proof?
    \item[] Answer: \answerNA{} 
    \item[] Justification: Not include theoretical results.

    \item {\bf Experimental result reproducibility}
    \item[] Question: Does the paper fully disclose all the information needed to reproduce the main experimental results of the paper to the extent that it affects the main claims and/or conclusions of the paper (regardless of whether the code and data are provided or not)?
    \item[] Answer: \answerYes{} 
    \item[] Justification: Detailed implementation details regarding model architectures, training hyperparameters, training pipelines, and training data are included in the main paper and Appendix.

\item {\bf Open access to data and code}
    \item[] Question: Does the paper provide open access to the data and code, with sufficient instructions to faithfully reproduce the main experimental results, as described in supplemental material?
    \item[] Answer: \answerYes{} 
    \item[] Justification: We have publicly released the code.

\item {\bf Experimental setting/details}
    \item[] Question: Does the paper specify all the training and test details (e.g., data splits, hyperparameters, how they were chosen, type of optimizer) necessary to understand the results?
    \item[] Answer:\answerYes{} 
    \item[] Justification: The main details are provided in the main paper and full details are provided in the Appendix.

\item {\bf Experiment statistical significance}
    \item[] Question: Does the paper report error bars suitably and correctly defined or other appropriate information about the statistical significance of the experiments?
    \item[] Answer: \answerYes{} 
    \item[] Justification: The Appendix provides statistical analysis for the core architecture. We repeat the key experiments five times with different random seeds and report the mean and standard deviation.

\item {\bf Experiments compute resources}
    \item[] Question: For each experiment, does the paper provide sufficient information on the computer resources (type of compute workers, memory, time of execution) needed to reproduce the experiments?
    \item[] Answer: \answerYes{} 
    \item[] Justification: We provide the compute resources in main paper.
    
\item {\bf Code of ethics}
    \item[] Question: Does the research conducted in the paper conform, in every respect, with the NeurIPS Code of Ethics \url{https://neurips.cc/public/EthicsGuidelines}?
    \item[] Answer: \answerYes{}{} 
    \item[] Justification: The research was conducted in full compliance with the NeurIPS Code of Ethics, adhering to established standards for responsible model development and fair experimental practices.

\item {\bf Broader impacts}
    \item[] Question: Does the paper discuss both potential positive societal impacts and negative societal impacts of the work performed?
    \item[] Answer: \answerYes{} 
    \item[] Justification: The paper discusses the potential positive impact of improving autonomous driving safety and efficiency through better planning. It also discusses limitations and failure cases.
    
\item {\bf Safeguards}
    \item[] Question: Does the paper describe safeguards that have been put in place for responsible release of data or models that have a high risk for misuse (e.g., pre-trained language models, image generators, or scraped datasets)?
    \item[] Answer: \answerNA{} 
    \item[] Justification:

\item {\bf Licenses for existing assets}
    \item[] Question: Are the creators or original owners of assets (e.g., code, data, models), used in the paper, properly credited and are the license and terms of use explicitly mentioned and properly respected?
    \item[] Answer: \answerYes{}{} 
    \item[] Justification: All third-party assets, including the datasets and pre-trained model are properly credited through citations and their original licenses have been reviewed and respected.

\item {\bf New assets}
    \item[] Question: Are new assets introduced in the paper well documented and is the documentation provided alongside the assets?
    \item[] Answer: \answerNA{} 
    \item[] Justification:

\item {\bf Crowdsourcing and research with human subjects}
    \item[] Question: For crowdsourcing experiments and research with human subjects, does the paper include the full text of instructions given to participants and screenshots, if applicable, as well as details about compensation (if any)? 
    \item[] Answer: \answerNA{} 
    \item[] Justification:

\item {\bf Institutional review board (IRB) approvals or equivalent for research with human subjects}
    \item[] Question: Does the paper describe potential risks incurred by study participants, whether such risks were disclosed to the subjects, and whether Institutional Review Board (IRB) approvals (or an equivalent approval/review based on the requirements of your country or institution) were obtained?
    \item[] Answer: \answerNA{} 
    \item[] Justification:

\item {\bf Declaration of LLM usage}
    \item[] Question: Does the paper describe the usage of LLMs if it is an important, original, or non-standard component of the core methods in this research? Note that if the LLM is used only for writing, editing, or formatting purposes and does \emph{not} impact the core methodology, scientific rigor, or originality of the research, declaration is not required.
    \item[] Answer: \answerNo{} 
    \item[] Justification: LLMs are not used as an important, original, or non-standard component of the core methodology.

\end{enumerate}
\end{document}